\RequirePackage{fix-cm}
\documentclass[twocolumn]{svjour3}           
\smartqed  
\usepackage{graphicx}
\usepackage[utf8]{inputenc}
\usepackage[english]{babel}
\usepackage{url}
\usepackage{color}
\usepackage{picinpar}  

\journalname{K\"unstliche Intelligenz}  
\begin{document}

\title{A System for Probabilistic Linking of Thesauri and Classification Systems
}


\author{Lisa Posch \and Philipp Schaer \and Arnim Bleier \and Markus Strohmaier}



\institute{Lisa Posch \at 
GESIS -- Leibniz Institute for the Social Sciences\\
Cologne, Germany \\
Institute for Web Science and Technologies\\
University of Koblenz-Landau, Germany \\
\email{lisa.posch@gesis.org}
\and
Philipp Schaer \at 
GESIS -- Leibniz Institute for the Social Sciences\\
Cologne, Germany \\
\email{philipp.schaer@gesis.org}
\and
Arnim Bleier \at 
GESIS -- Leibniz Institute for the Social Sciences\\
Cologne, Germany \\
\email{arnim.bleier@gesis.org}
\and
Markus Strohmaier \at 
GESIS -- Leibniz Institute for the Social Sciences\\
Cologne, Germany \\
Institute for Web Science and Technologies\\
University of Koblenz-Landau, Germany \\
\email{markus.strohmaier@gesis.org}
}

\date{}

\maketitle

\begin{abstract}
This paper presents a system which creates and visualizes probabilistic semantic links between concepts in a thesaurus and classes in a classification system. For creating the links, we build on the \emph{Polylingual Labeled Topic Model (PLL-TM)} \cite{plltm2015}. PLL-TM identifies probable thesaurus descriptors for each class in the classification system by using information from the natural language text of documents, their assigned thesaurus descriptors and their designated classes.
The links are then presented to users of the system in an interactive visualization, providing them with an automatically generated overview of the relations between the thesaurus and the classification system.

\end{abstract}

\section{Introduction}
\label{intro}
Thesauri and classification systems are two different kinds of \emph{Knowledge Organisation Systems (KOS)} which are widely used in Library and Information Science for annotating documents and other artifacts in (digital) libraries, archives or information systems \cite{weller2010knowledge}. While both KOS share the common foundational idea of using structured controlled vocabularies to describe concepts, they differ in terms of complexity and broadness. 

Differing structural features of thesauri and classification systems render them suitable for different types of applications, as pointed out by Tudhope et al.:
``While the structure of a classification system
and a thesaurus may be fairly similar, in that both consist of hierarchical structures of concepts, they will tend to differ in the \emph{exhaustivity} and \emph{specificity} of their application to information items. Thus an information item will generally tend to be classified by fewer, more general concepts from a classification system and conversely will tend to be indexed by several, more specific concepts from a thesaurus." \cite[p. 25]{bath23563}

Consequently, many digital libraries\footnote{Like PubMed, ACM Digital Library, RePeC, EconStor, SSOAR, and others.} use different KOS to describe each document in order to complement their advantages. The usage of different KOS is helpful, for example, in an interactive retrieval scenario where users can drill down on thesaurus- or keyword-based search results by using a classification-based filter. Another example of hybrid use of thesauri and classification systems is search term recommendation, where specialised recommender systems were trained for different sub-classifications of a text corpus \cite{lueke2012}.
While both classification systems and thesauri have semantic links between the concepts which they contain, semantic links between the two KOS are missing.

\begin{figure}[t!]
		\includegraphics[width = 1\columnwidth]{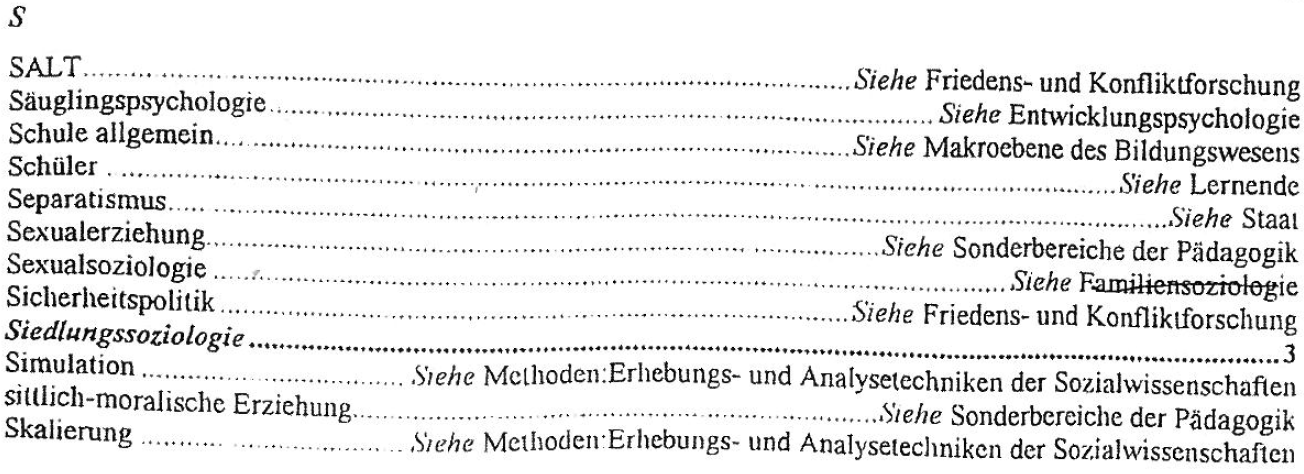}
	\caption{\textbf{Excerpt of a ``cheat sheet".} This figure shows an excerpt of a ``cheat sheet", a document offering informal descriptions of possible links between a thesaurus and a classification system. These documents are created in order to support the work of human indexers who work simultaneously with both KOS.}
	\label{fig:cheatsheet}
\end{figure}

\begin{figure*}[t!h]
\centering
		\includegraphics[width = 0.82\textwidth]{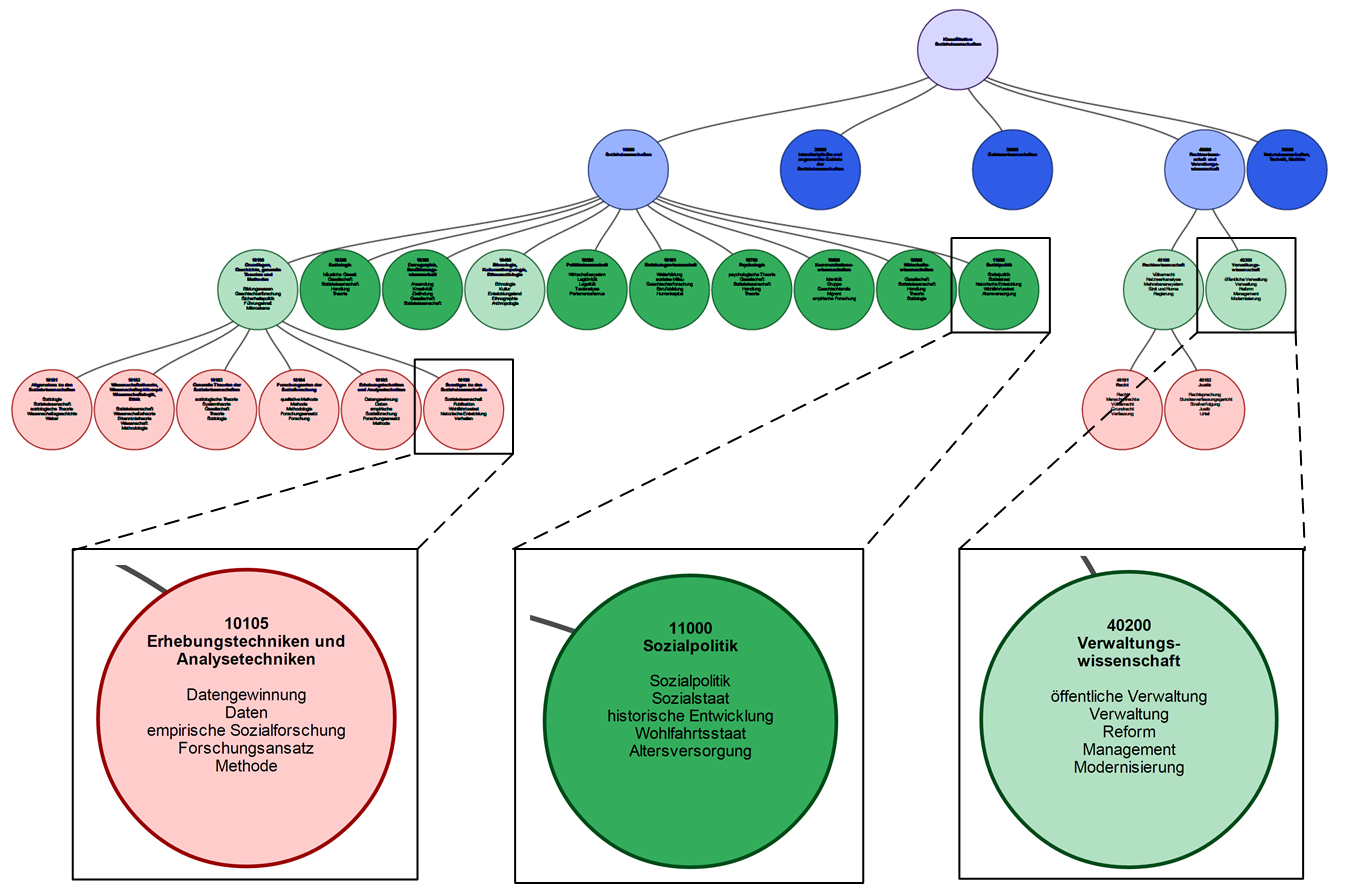}
	\caption{\textbf{Visualization of the links.} This figure shows a screenshot of the system's user interface as well as a zoomed-in close-up of three links. Each link between a class in the classification system and thesaurus descriptors is represented by a tree node. The tree nodes contain the class number, the name of the class and the five most probable thesaurus descriptors inferred by the system through PLL-TM.
Sub-trees are interactive and collapsible/expandable so that the user only sees those parts of the classification system which he or she is interested in. Dark nodes have children (are expandable), light nodes are leaves or already expanded nodes.}
	\label{fig:the_system}
\end{figure*}

\textbf{Problem.} In manual and semi-automatic indexing, human indexers work simultaneously with both KOS. 
Currently, the only support for displaying a link between the two KOS is offered in form of a manually created ``cheat sheet", a document listing rules of thumb for selecting a class from the classification system, once a thesaurus concept has been chosen.
However, manual creation of such links is a labor-intensive process and therefore, ``cheat sheets" often merely provide incomplete and outdated support to users.
Figure \ref{fig:cheatsheet} depicts an excerpt of such a document.
To address this shortcoming, we present a system which computes the latent semantic links between thesauri and classification systems and visualizes them in an explicit way. 

\textbf{Contribution.} The main contribution of this paper is a system which creates and visualizes probabilistic semantic links between thesaurus descriptors and classes contained in a classification system. These links are generated by the \emph{Polylingual Labeled Topic Model (PLL-TM)} \cite{plltm2015} from document corpora where both KOS are used simultaneously in the metadata curation. The probabilistic links are then visualized in an interactive way. 
While these semantic links were previously only available in a manual ad-hoc or informal manner, we can now systematically identify and visualize these links.
To the best of our knowledge, this is the first system which uses topic models for identifying latent semantic links between different KOS of the same domain.

\section{System Description}

The system consists of two main components:
The first component is responsible for extracting the links with PLL-TM, whereby each class in the classification system is described by a distribution over thesaurus descriptors. 
The second component is an interactive brow\-ser-based visualization of the links, implemented using the Java\-Script library D3\footnote{http://d3js.org/}.
Here, the links, i.e. the classes contained in the classification system along with their most probable thesaurus descriptors, are visualized as nodes in an interactive collapsible tree. 

\textbf{Probabilistic Link Extraction.} PLL-TM is a supervised topic model which accounts for multiple languages while simultaneously restricting the topics of a document to its labels. In the context of this system, PLL-TM models two languages in order to create links between classes and descriptors: the natural language words of the documents represent the first language, and the controlled vocabulary of the thesaurus constitute the second language. 
The classes with which a document is annotated represent the documents' labels.

This setup gives the system the ability to probabilistically link thesaurus descriptors to classes in a classification system, using both the natural language text of the abstracts and the assigned thesaurus descriptors as information for creating the links. 
PLL-TM then generates one topic per class in the classification system, described by a distribution over thesaurus descriptors.

\textbf{Interactive Visualization.} The second component takes the links generated by PLL-TM as input and displays them in an interactive browser-based visualization. Classification systems have a hierarchical tree-structure, which is kept in the visualization.
Each link is displayed as one node in the classification tree, containing the class that it represents as well as the five most probable thesaurus descriptors for this class. 
Figure \ref{fig:the_system} shows a screenshot of the system applied to linking a thesaurus and a classification system from the social sciences domain (also see Section \ref{sec:application}) as well as a zoomed-in close-up of three nodes, i.e. links.
Sub-trees of the classification system are interactively expandable and collapsible so that only those parts are displayed that the user is interested in. In the visualization, the nodes which are expandable (i.e. have undisplayed children) are displayed in a dark color, the nodes which are already expanded or are leaves are displayed in a lighter color. The implementation of the visualization uses D3, a JavaScript library for creating data-driven documents \cite{bostock2011d3}. The arrangement of the nodes is based on Bostock's implementation \cite{treeimpl} 
of the Reingold-Tilford algorithm \cite{reingold1981tidier}.

\section{Application of the System in the Social Science Domain}
\label{sec:application}

We apply the system to linking two KOS from the social sciences domain: the SKOS \emph{Thesaurus for the Social Sciences (TheSoz)} \cite{zapilko2013thesoz} and the \emph{Classification for the Social Sciences (CSS)}.
The CSS is used for disciplinary categorization of documents and consists of 159 classes in four hierarchy levels.
TheSoz is used for subject indexing and contains about 8.000 descriptors as well as 4.000 non-descriptors (i.e. synonyms).
For extracting the links, we use documents from the \emph{Social Science Literature Information System (SOLIS)}.
SOLIS contains metadata of German social science publications, and all publications have been manually indexed with the TheSoz and manually classified with the CSS by human domain experts.
However, currently, the two KOS are not linked beyond their joint usage for individual documents.

The system's first component trains PLL-TM on the abstracts and TheSoz descriptors of SOLIS documents, restricting the permitted topics for each document to its assigned classes. In our case, we used all documents which were published in the years 2008 to 2013, resulting in a corpus of about 60.000 documents.
The second component visualizes the links: For example, the class \emph{``40200 Administrative Science"} is a second-level class in the CSS hierarchy which the system linked to the descriptors \emph{public administration}, \emph{administration}, \emph{reform},  \emph{management} and \emph{modernization} (translation of the rightmost close-up in Figure \ref{fig:the_system}).

\textbf{Usage Scenarios.} The context in which our system will be applied is the classification and semantic indexing process in the social science domain.
Working with domain-specific KOS is a non-trivial issue in every complex scientific domain, but this is especially prominent in the social sciences where there is a high heterogeneity in the use of specialized languages \cite{petras2006translating} in different sub-domains. Therefore, a significant amount of domain knowledge and experience is usually required to work with the KOS. 
A visualization of the latent semantic links between two KOS can help mitigate this complexity for different groups of users of repositories or digital library systems: (1) trainee indexers who just started to learn to work with KOS, (2) authors who want to classify their publications, (3) researchers who want to improve the retrieval process by incorporating knowledge about the structure and meaning of concepts in the KOS, or (4) expert indexers who are already familiar with the KOS but will benefit from a structured visualization of the latent semantic links between two KOS.

While numerous different applications of the system are conceivable, we highlight one specific use case: the system's use during the subject indexing process.
In the first step of this process, one or more classes are selected for a new document, either predicted automatically or chosen manually.
In the case of automatic class prediction, we have already shown that using SVM or topic models to predict the top hierarchy levels of the classification leads to good results \cite{posch2014}.
If the class is selected manually, the linked thesaurus descriptors facilitate the decision by serving as descriptions of the concepts which the classes represent. 
Then, once a class is chosen for a document, the most probable descriptors for this class provide the user with a set of likely annotation candidates to choose from.
This two step approach is useful due to the lower dimensionality of the classification system compared to the thesaurus: Intuitively, choosing one class out of 159 is easier than choosing one descriptor out of 8.000.

\section{Conclusion}

In this paper, we presented a system for probabilistic semantic linking of thesaurus descriptors to classes in a classification system. 
The system consists of two components: (1) probabilistic semantic link generation using PLL-TM and (2) interactive visualization of the links. Identifying and visualizing latent semantic links between two KOS is helpful in scenarios such as exploratory classification browsing and as support during the subject indexing process.

For future work, we plan to apply the system in different domains and evaluate the extracted links with human expert indexers. Furthermore, we plan to develop a (semi-)automatic annotation system which exploits the links not only at the global KOS level, but suggests classes and a set of thesaurus descriptors for individual documents.



\bibliographystyle{spmpsci}      
\bibliography{biblio}   

%
%

\end{document}